\def\year{2022}\relax
\documentclass[letterpaper]{article} % DO NOT CHANGE THIS
\usepackage{aaai22}  % DO NOT CHANGE THIS
\usepackage{times}  % DO NOT CHANGE THIS
\usepackage{helvet}  % DO NOT CHANGE THIS
\usepackage{courier}  % DO NOT CHANGE THIS
\usepackage[hyphens]{url}  % DO NOT CHANGE THIS
\usepackage{graphicx} % DO NOT CHANGE THIS
\usepackage{natbib}  % DO NOT CHANGE THIS AND DO NOT ADD ANY OPTIONS TO IT
\usepackage{caption} % DO NOT CHANGE THIS AND DO NOT ADD ANY OPTIONS TO IT
\DeclareCaptionStyle{ruled}{labelfont=normalfont,labelsep=colon,strut=off} % DO NOT CHANGE THIS
\usepackage{amsmath}
\usepackage{amssymb}
\usepackage{multirow}
\usepackage{threeparttable}
\usepackage{subfigure}
\usepackage{dsfont}

\title{Unsupervised Domain Adaptive Salient Object Detection Through Uncertainty-Aware Pseudo-Label Learning}
\author{
    Pengxiang Yan\textsuperscript{\rm 1,2\thanks{The first two authors have equal contribution. $^\dag$Corresponding author is Guanbin Li (Email: liguanbin@mail.sysu.edu.cn).}},
    Ziyi Wu\textsuperscript{\rm 1\footnotemark[1]},
    Mengmeng Liu\textsuperscript{\rm 1},
    Kun Zeng\textsuperscript{\rm 1},
    Liang Lin\textsuperscript{\rm 1},
    Guanbin Li\textsuperscript{\rm 1,3\footnotemark[2]}
}
\affiliations{
    \textsuperscript{\rm 1}Sun Yat-sen University
    \textsuperscript{\rm 2}ByteDance Inc.
    \textsuperscript{\rm 3}Shenzhen Research Institute of Big Data\\
    yanpx@live.com, wuzy39@mail2.sysu.edu.cn, liumm97@outlook.com, \\
    zengkun2@mail.sysu.edu.cn, linliang@ieee.org, liguanbin@mail.sysu.edu.cn
}

\begin{document}

\maketitle

\begin{abstract}
Recent advances in deep learning significantly boost the performance of salient object detection (SOD) at the expense of labeling larger-scale per-pixel annotations. To relieve the burden of labor-intensive labeling, deep unsupervised SOD methods have been proposed to exploit noisy labels generated by handcrafted saliency methods. However, it is still difficult to learn accurate saliency details from rough noisy labels. In this paper, we propose to learn saliency from synthetic but clean labels, which naturally has higher pixel-labeling quality without the effort of manual annotations. %is much more easy to obtain. 
Specifically, we first construct a novel synthetic SOD dataset by a simple copy-paste strategy.
Considering the large appearance differences between the synthetic and real-world scenarios, directly training with synthetic data will lead to performance degradation on real-world scenarios. To mitigate this problem, we propose a novel unsupervised domain adaptive SOD method to adapt between these two domains by uncertainty-aware self-training. 
% Specifically, it performs iterative training with both synthetic labels and real-world pseudo-labels. After each training iteration, the pseudo-labels will be updated through an uncertainty-aware pseudo-label learning strategy that includes image-level sample selection and pixel-wise pseudo-label reweighting.
Experimental results show that our proposed method outperforms the existing state-of-the-art deep unsupervised SOD methods on several benchmark datasets, and is even comparable to fully-supervised ones.
\end{abstract}

%%%%%%%%% BODY TEXT
\section{Introduction}
Salient object detection (SOD) aims to accurately locate and segment out the most visually distinctive object region in a scene.
%As it can naturally narrow down the scope of visual processing to reduce computational costs, it has severed as an important pre-processing step and been applied to various computer vision tasks, such as video compression~\cite{hadizadeh2013saliency}, image retrieval~\cite{gao2015database} and %saliency-aware video object segmentation~\cite{wang2017saliency}, photo cropping~\cite{wang2018deep}, \textit{etc}.
\begin{figure}[t]
    \centerline{
       \includegraphics[width=0.47\textwidth]{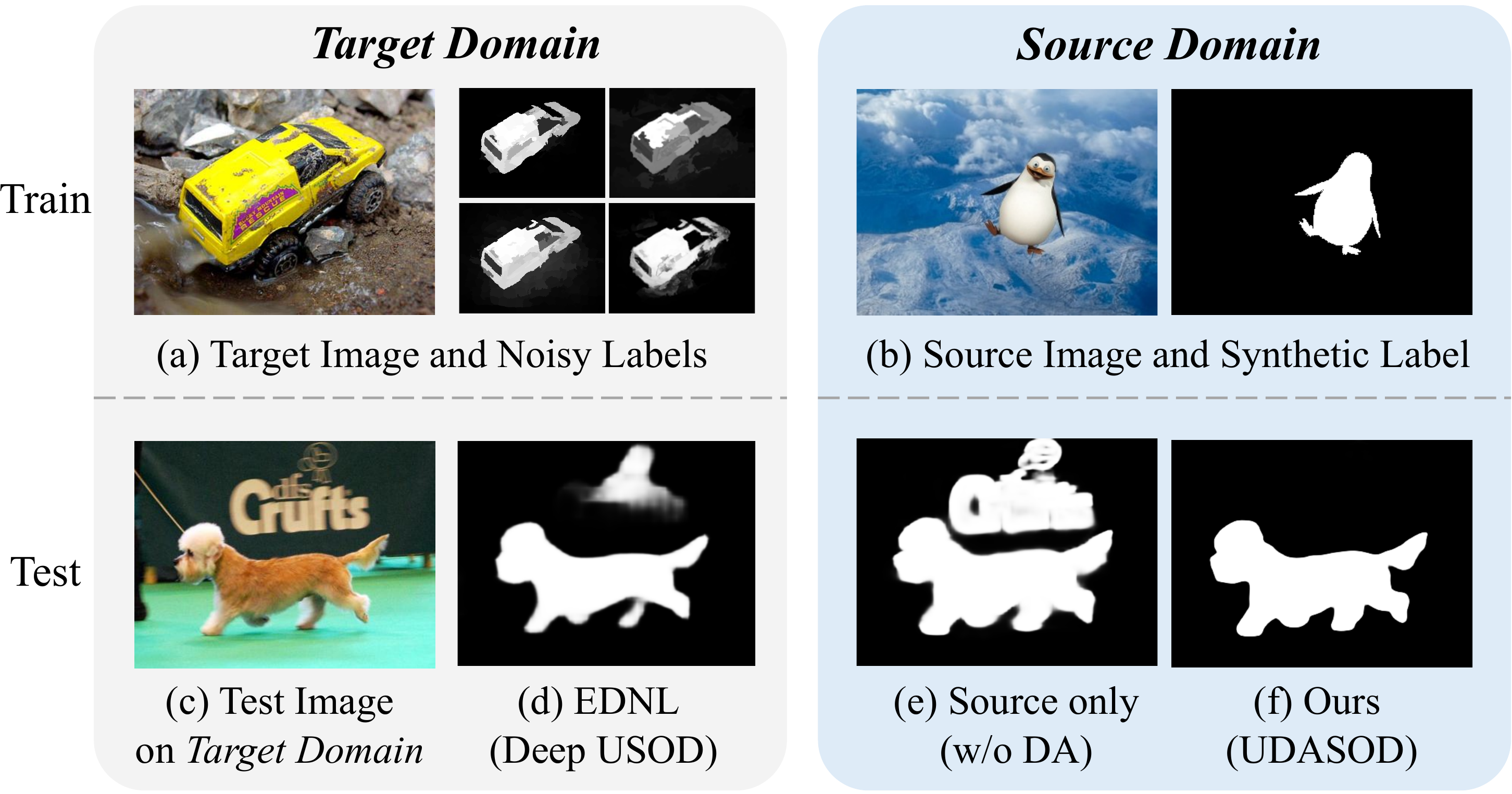}
    }
    \caption{Deep unsupervised salient object detection (USOD) achieved by two training settings. Existing deep USOD algorithms are mainly trained on \textbf{(a)} real-world images (target domain) with noisy labels generated by traditional USOD methods. While we propose to exploit \textbf{(b)} the synthetic saliency data (source domain) for training. However, due to the discrepancy between two domains, the saliency detector trained only on synthetic data (\textbf{(e)} Source only) without domain adaption (DA) usually fails to performs well on \textbf{(c)} real images. To solve this problem, we propose \textbf{(f)} an unsupervised domain adaptive SOD (UDASOD) method, which can generate more accurate saliency predictions than \textbf{(d)} the best-performing deep USOD method EDNL~\cite{zhang2020learning}.}
    \label{fig:intro}
\end{figure}
In recent years, the development of deep convolutional neural networks (DCNN) significantly boosts the performance of salient object detection%. The DCNN-based SOD models can be trained end-to-end to produce saliency maps of high accuracy. Thus, the DCNN-based algorithms
~\cite{wei2020label,qin2019basnet,wang2018detect} and has taken place of conventional hand-crafted feature-based algorithms ~\cite{zhang2015minimum,zhu2014saliency,li2013saliency} to become the dominant methods in salient object detection. However, such promising performance comes at a cost of a large number of pixel-wise annotated images to train the DCNN-based models. Moreover, to ensure the quality and consistency of labeling, it generally requires multiple human annotators to annotate fine pixel-level masks for the same image~\cite{li2017instance,fan2018salient}. The time-consuming and laborious labeling work limits the amount of training data and thus hampers the further development of DCNN-based SOD methods.

To alleviate the burden of pixel-wise labeling but take full advantage of the end-to-end training advantages of DCNN, weakly-supervised~\cite{zhang2020weakly,zeng2019multi,li2018weakly} and deep unsupervised~\cite{zhang2020learning,nguyen2019deepusps,zhang2018deep} SOD algorithms have been proposed. Weakly-supervised SOD algorithms mainly focus on learning saliency inference from simple but clean manual annotations, such as image classes~\cite{li2018weakly}, image captions~\cite{zeng2019multi}, and scribbles~\cite{zhang2020weakly}. 
While deep unsupervised SOD methods aim to learn saliency detection without resorting to any manual annotations. 
Existing deep unsupervised SOD methods mainly focus on learning from the dense noisy labels generated by single~\cite{zhang2020learning_pami} or multiple~\cite{zhang2020learning,nguyen2019deepusps,zhang2018deep,zhang2017supervision} traditional unsupervised SOD methods (as shown in Fig.~\ref{fig:intro}~(a) ), which can be achieved through noise modeling~\cite{zhang2020learning,zhang2018deep} or pseudo-label self-training~\cite{zhang2020learning_pami,nguyen2019deepusps,zhang2017supervision}. However, traditional unsupervised methods that rely on manual features and specific saliency priors are arduous to deal with the complex situation of low foreground/background contrast. The generated pseudo-labels are rich in noise and are almost impossible to be repaired in iterative training based on pseudo-labels, especially for the boundary of the salient objects. 
%However, the noisy labels are still too rough for saliency detectors to learn accurate saliency information especially around object boundaries, which be the bottleneck of the further development of deep unsupervised SOD methods. 

Instead of struggling with the generated noisy labels of real images, in this paper, we propose that learning saliency from the synthetic but clean labels (Fig.~\ref{fig:intro}~(b)) would be yet another feasible solution. There are massive object images with transparent backgrounds as well as pure background images without salient objects that can be easily collected from the design resources or photography websites on the Internet. Since the salient objects of a scene are usually the foreground objects, we construct a new large-scale synthetic salient object detection (SYNSOD) dataset with clean labels by simply copying foreground objects and pasting them on the background images. The SYNSOD dataset can be applied to existing fully-supervised SOD methods to relieve the burden of manual annotations. However, as shown in Fig.~\ref{fig:intro},  due to the presence of large appearance differences between real images (\textit{target domain}) and synthetic images (\textit{source domain}) known as ``domain gap'', the model directly trained on SYNSOD (Fig.~\ref{fig:intro} (e)) 
fails to performs well on the real-world dataset such as DUTS~\cite{wang2017learning}.

To resolve the above issues, we propose a novel unsupervised domain adaptive salient object detection (UDASOD) algorithm to adapt the DCNN-based saliency detector trained on the synthetic dataset to the real-world SOD datasets. % in an unsupervised manner without resorting to any manual annotations. 
The proposed UDASOD algorithm is an iterative method that exploits an uncertainty-aware pseudo-learning (UPL) strategy to achieve adaption between two domains. Specifically, in each round of iteration, UDASOD leverages the source images with synthetic labels and the target images with weighted pseudo-labels to jointly train the saliency detector. After the training of each round, UPL dynamically updates the training set and pseudo-labels of the target domain through three major steps, including pixel-wise uncertainty estimation, image-level sample selection and pixel-wise pseudo-label reweighting. %Firstly, it estimates the pixel-wise uncertainty of each target sample based on the consistency between the generated pseudo-label and the resulting saliency maps of the target images processed with different data augmentation. Secondly, it ranks the target samples according to an image-level uncertainty of pseudo-labels and selects the samples with low uncertainty forming as a new target training set. Thirdly, it performs pixel-wise pseudo-label reweighing to assign different loss weights to pseudo-labels according to their pixel-wise uncertainty maps. 
%Finally, the newly selected target training set and weighted pseudo-labels will be fed to the training of the next round. 
The main contributions of this paper can be summarized as follows:
\begin{itemize}
    \item To our knowledge, we are the first attempt to achieve SOD by exploiting unsupervised domain adaption from synthetic data, which varies from existing deep unsupervised SOD algorithms targeted at noisy labels. 
    \item We construct a synthetic SOD dataset %without resorting to any manual annotations. We
    and further propose UDASOD that exploits uncertainty-aware pseudo-label learning to adapt the saliency detector trained on the synthetic dataset to real-world scenarios.
    \item Experimental results show that our proposed domain adaptive SOD method outperforms all existing state-of-the-art deep unsupervised SOD methods and is comparable to the fully-supervised ones.
\end{itemize}
%------------------------------------------------------------------------
\section{Related Work}
\subsection{Salient Object Detection}
Conventional SOD is mainly achieved by different saliency priors or handcrafted features~\cite{zhang2015minimum,zhu2014saliency,li2013saliency}. Recent advances in DCNNs significantly boost the performance of SOD at a cost of numerous pixel-wise annotations \cite{chen2020global,wei2020f3net,pang2020multi,liu2019employing,wang2018detect}.
%with end-to-end learning paradigms. 
%Those fully-supervised methods mainly focus on designing different network architectures to promote the saliency representation~\cite{pang2020multi,liu2019employing,wang2018detect}. 
To mitigate the labeling costs, weakly supervised SOD is proposed to learn saliency under weak supervision such as image classes~\cite{li2018weakly}, captions~\cite{zeng2019multi}, and scribbles~\cite{zhang2020weakly}. Deep unsupervised SOD is further proposed to learn saliency without resorting to any manual annotations. Existing deep unsupervised methods mainly rely on learning from noise labels generated by conventional SOD methods, which can be achieved through noise modeling~\cite{zhang2020learning,zhang2018deep} or pseudo-label self-training~\cite{zhang2020learning_pami,nguyen2019deepusps}. In this paper, we propose to solve SOD from a novel perspective, \textit{i.e.}, learning from synthetic but clean labels.

\subsection{Unsupervised Domain Adaption}
Unsupervised domain adaptation (UDA) aims to transfer the knowledge learned from the label-rich source domain to an unlabeled target domain. It is widely studied on various vision tasks such as image classification~\cite{sener2016learning}, object detection~\cite{chen2018domain}, semantic segmentation~\cite{chen2017no}, \textit{etc}. Among these tasks, semantic segmentation shares most characteristics with SOD. The primary approach of UDA for semantic segmentation is to minimize the discrepancy between two domain distributions through adversarial learning~\cite{hoffman2018cycada,luo2019significance,tsai2018learning}. There are some self-training-based UDA methods~\cite{zou2018unsupervised, zou2019confidence} that assign pseudo-labels to confident target samples and directly use pseudo-labels as target domain supervision to reduce domain mismatch.
%In this paper, we exploit the self-training-based UDA to alleviate the domain discrepancy between the synthetic and real salient object detection datasets. 
To the best of our knowledge, we are the first attempt to relieve the burden of large-scale manual annotations by leveraging UDA on salient object detection.
%apply UDA to unsupervised salient object detection.

\subsection{Pseudo-Label Learning}

Pseudo-label learning, which is initially explored in semi-supervised learning scenario~\cite{lee2013pseudo}, has recently attracted wide attention due to its simplicity and effectiveness. The goal of pseudo-label learning is to fully exploit the unlabeled data by generating and updating pseudo-labels for unlabeled samples with a model trained on labeled data. Thus, it can be applied to benefit various tasks such as semi-supervised learning~\cite{lee2013pseudo,yan2019semi}, domain adaption~\cite{zheng2021rectifying,li2020content}, and noisy label learning~\cite{zhang2020learning_pami,tanaka2018joint}. There are also some SOD methods \cite{li2018weakly,nguyen2019deepusps} that exploit the pseudo-label learning technique. 
%Li \textit{et al.} ~\cite{li2018weakly} exploit class activation maps refined with fully-connected CRF~\cite{krahenbuhl2012efficient} to update pseudo-labels. Nguyen \textit{et al.}~\cite{nguyen2019deepusps} exploit pseudo-label learning based on multiple noise labels and iteratively refine pseudo-labels with moving average and fully-connected CRF. 
Different from them, our proposed method exploits an uncertainty-aware pseudo-learning strategy that treats each pseudo-label differently and is also free of time-consuming post-processing like fully-connected CRF.

%------------------------------------------------------------------------

\begin{figure}[t]
    \centerline{
      \includegraphics[width=0.47\textwidth]{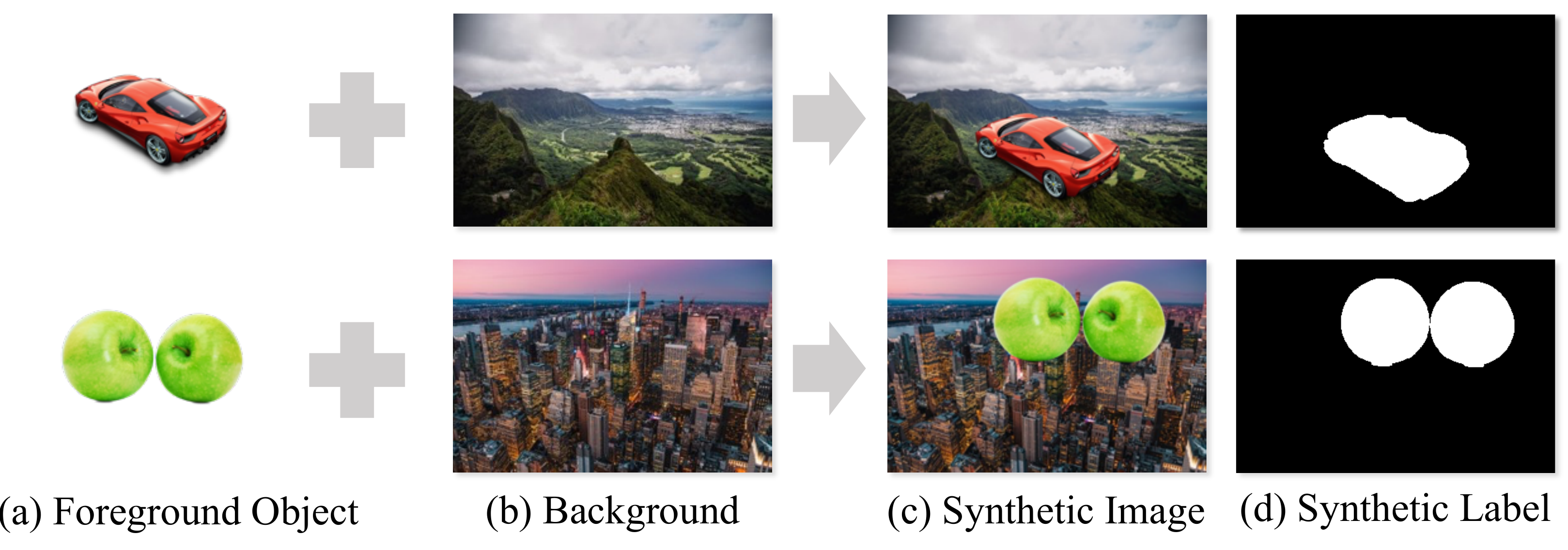}
    }
    \caption{Examples of the dataset construction of SYNSOD. Each foreground object is matched with a unique background to generate a synthetic sample through a simple copy-paste strategy. The pixel-level synthetic label can be obtained from the alpha channel of the foreground image.}
    \label{fig:dataset_example}
\end{figure}

\section{Proposed Dataset}
In this section, we detail the proposed SYNSOD dataset from the following aspects.

%\subsection{Dataset Construction.}
\textbf{Image Collection.} As salient objects are usually the foreground objects of a scene, we can intuitively obtain a synthetic image with salient objects by pasting the foreground objects on a background image. Thus, to construct a novel synthetic SOD dataset, we first collect a large number of object images with transparent backgrounds (RGBA color) from several websites with non-copyrighted design resources, each of which contains single or multiple objects of diverse appearances and categories. Next, we collect background photos from multiple non-copyrighted photography websites, which contains various non-salient scenes such as forest, grass, sky, ocean, \textit{etc}. The collection process is executed through a designed spider program and images with low resolution will be automatically removed.

\textbf{Data Generation.} Given the premise of the collected foreground and background images, we can easily generate the synthetic SOD dataset by a simple copy-paste strategy. As shown in Fig.~\ref{fig:dataset_example}, we match each foreground object image with a unique non-salient background image. Then, we randomly scale the object image with a ratio ranging from 0.5 to 1.1. Next, we set an object center in the background image to cover its surrounding pixels with the non-transparent object pixels, resulting in the synthetic image for SOD. The pixel-level synthetic label can be easily obtained by binarizing the corresponding alpha channel of the foreground object pixels in a synthetic image. %To match the distribution of the real-scene salient dataset, 40\% of object centers are set in the image center and the rest are randomly located. 
In this way, we construct a large-scale synthetic SOD dataset (SYNSOD), containing 11,197 synthetic images and corresponding pixel-level labels.

\begin{figure}[t]
    \centerline{
       \includegraphics[width=0.5\textwidth]{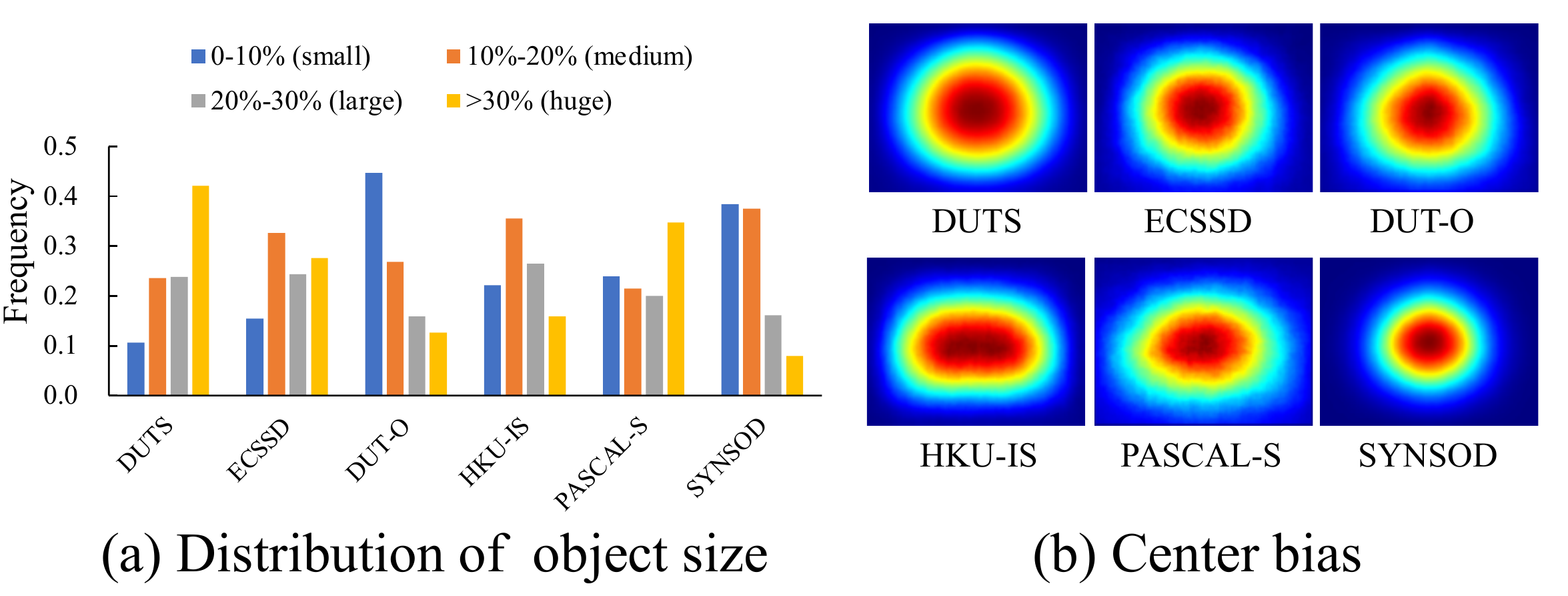}
    }
    \caption{Statistics of our proposed SYNSOD dataset including  distribution of salient object size and center bias.}
    \label{fig:dataset_statistic}
\end{figure}

\textbf{Dataset Statistics.} As shown in Fig.~\ref{fig:dataset_statistic},
 we present the following dataset statistics on our proposed SYNSOD dataset and five public benchmark SOD datasets~\cite{wang2017learning,yan2013hierarchical,yang2013saliency,li2015visual,li2014secrets}. \textbf{1) Object size.} As shown in Fig.~\ref{fig:dataset_statistic} (a), the ratio of salient object size in SYNSOD ranges from 0.39\% to 86.96\% (avg.: 14.72\%), yielding a border range.
\textbf{2) Center bias.} To reveal the degree of center bias, we compute the average saliency maps over all images of each dataset. As shown in Fig.~\ref{fig:dataset_statistic} (b), SYNSOD is center-biased and the degree of center-bias is slightly stronger than others, %which can be ascribed to the setting in the image synthesis process.
which shows strong domain gaps with other real-world datasets.

%------------------------------------------------------------------------

\begin{figure*}[ht]
    \centerline{
      \includegraphics[width=0.92\textwidth]{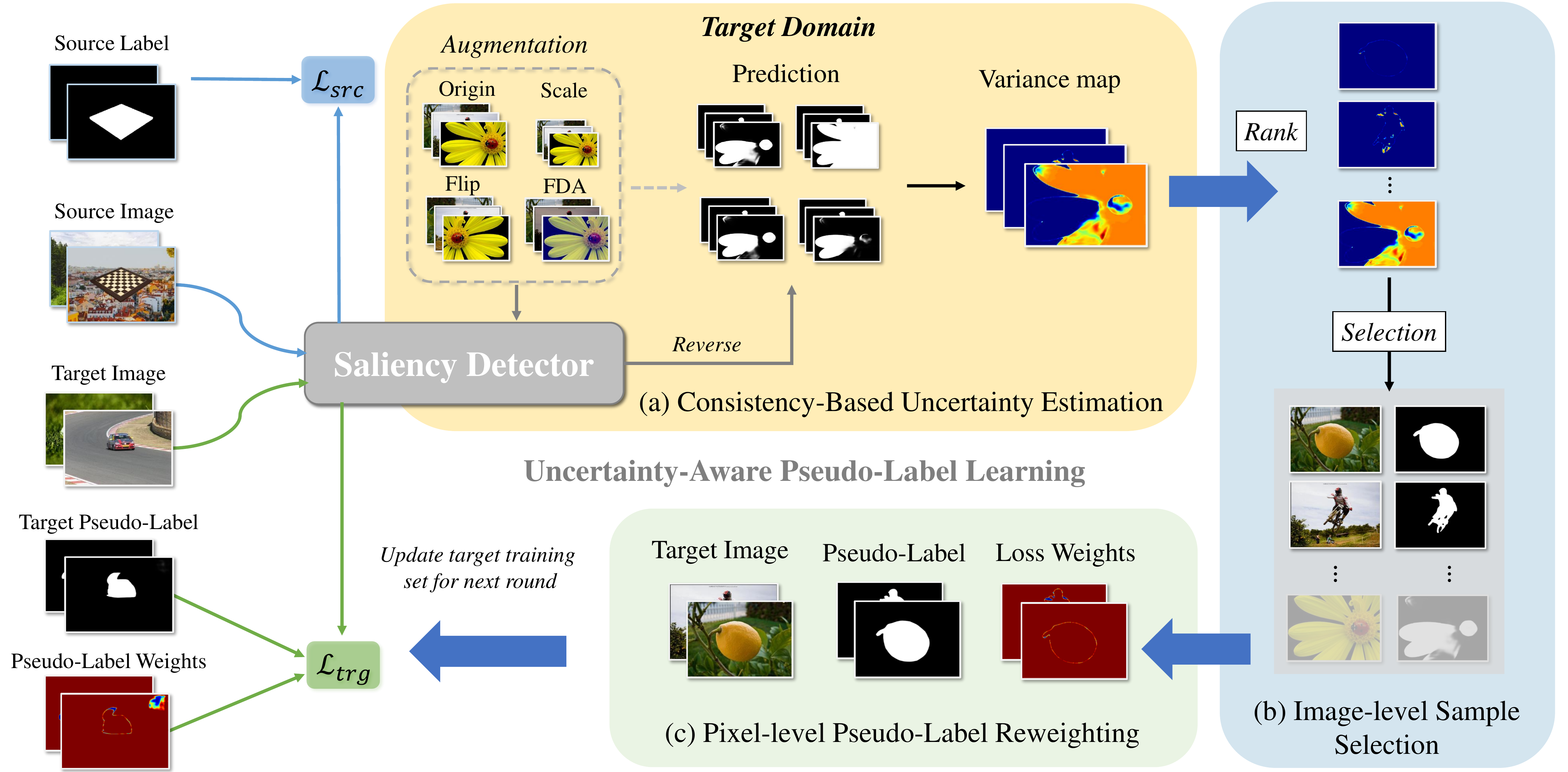}
    }
    \caption{The overall framework of our proposed unsupervised domain adaptive salient object detection method. It iteratively learns saliency from the synthetic labels  (source domain) and pseudo-labels of real images (target domain). The pseudo-labels will be dynamically updated after each training round through an uncertainty-aware pseudo-label learning strategy that contains three major steps, \textit{i.e.}, (a) consistency-based uncertainty estimation, (b) image-level sample selection, and (c) pixel-wise pseudo-label reweighting. We use three kinds of data augmentations for consistency-based uncertainty estimation, including 1) horizontal flipping (Flip), 2) rescale input image to $224\times224$ (Scale), and 3) randomly swap image style with other target images via FDA~\cite{yang2020fda}.}
    \label{fig:pipeline}
\end{figure*}
\section{Methodology}
%\subsection{Unsupervised Salient Object Detection}
\subsection{Problem Formulation}
To achieve SOD without resorting to manual annotations or noisy labels, we propose to learn saliency from the synthetic but clean labels through a novel unsupervised domain adaptive salient object detection (UDASOD) framework. As shown in Fig.~\ref{fig:pipeline}, UDASOD is formulated as an iterative training paradigm, which can leverage existing deep learning-based saliency detectors to learn saliency prediction from synthetic source data and unsupervisedly adapt it to the real target scenarios. To fully exploit unlabeled target images, UDASOD is jointly trained with the pseudo-labels of target images and the synthetic labels of source images. 
To formulate UDASOD, we start with the synthetic training set denoted as the source domain $\mathcal{D}_{src}=\{(I_s,y_s)\}_{s=1}^{S^i}$, where $I_s$ is a synthetic RGB color image of size $H\times W$, $y_s \in \{0, 1\}^{H\times W}$ is the corresponding binary saliency map, and $S^i$ is the number of source images in round $i$. The proposed UDASOD framework will unsupervisedly adapt the saliency detector from the synthetic dataset to the real SOD dataset denoted as target domain $\mathcal{D}_{trg}=\{(I_t,\hat y_t)\}_{t=1}^{T^i}$, where $I_t$ is a real RGB color image, $ \hat y_t\in[0,1]^{H\times W}$ is the corresponding pseudo-label, and $T^i$ is the number of pseudo-labels in round $i$. Thus, the training process of round $i$ can be formulated as optimization of the network parameters $\theta$ of the saliency detector as follows:
\begin{equation}
    \theta^i=\arg\min_\theta\mathcal{L}(\theta,i),
\end{equation}
where the loss function $\mathcal{L}(\theta, i)$ under the joint supervision of source $\mathcal{D}_{src}$ and target $\mathcal{D}_{trg}$  domains is defined as:
\begin{equation}
    \mathcal{L}(\theta | i)= \mathcal{L}_{src}(\theta|I_s^i,Y_s^i)+\mathcal{L}_{trg}(\theta|I_t^i, \hat 
 Y_t^i).
\end{equation}
Here, $Y_s$ and $\hat Y_t$ denote the set of synthetic source labels and the set of pseudo target labels, respectively. $\mathcal{L}_{src}$ and $\mathcal{L}_{trg}$ refer to the specific loss calculation of source and target samples, which will be detailed in the following.   

However, since the pseudo-labels of target domain are generated by the saliency detector initially trained on the source domain, the pseudo-labels inevitably contain incorrect pixel-level prediction due to the significant distribution gap between the two domains. To avoid error accumulation in the iterative training process, we propose that the samples of the target domain need to be carefully selected to participate in the training, and each pixel of the  selected sample should be adaptively assigned different weights. Therefore, the loss function for each predicted saliency map  $p\in[0,1]^{H\times W}$ is formulated with a weight matrix $\omega\in(0,1]^{H\times W}$ as follows:
\begin{equation}
    \mathcal{L}(y,p,\omega)=\sum_{h=1}^H\sum _{w=1}^W\omega^{(h,w)}\ell(y^{(h,w)},p^{(h,w)}),
\end{equation}
where $\ell(.)$ denotes the binary cross-entropy loss for each pixel and $y\in[0,1]^{H\times W}$ denotes the dense label of $p$. Then, the loss function for the source and target samples can be formulated as:
\begin{gather}
    \mathcal{L}_{src}(\theta|X_s^i,Y_s^i)=\sum^{S_i}_{s=1}     \mathcal{L}(y_s,p_\theta(I_s),\omega_s), \\
    \mathcal{L}_{trg}(\theta|X_t^i, Y_t^i) = \sum^{T_i}_{t=1}\mathcal{L}(\hat y_t, p_\theta(I_t), \omega_t), 
    \label{eq:loss_trg}
\end{gather}
where $p_\theta(I)$ denotes the prediction of saliency detector with parameters $\theta$ for input image $I$. In practice, we only assign different pixel-wise weights to pseudo-labels while setting $\omega_s = \mathds{1} \in \mathbb{R}^{H\times W}$ in source domain. At the end of each round, the target training set with pseudo-labels will be dynamically updated and assigned with pixel-level weights based on our proposed uncertainty-aware pseudo-label learning strategy. 

\subsection{Uncertainty-Aware Pseudo-Label Learning}
\label{sec:upl}
Instead of equally using all the pseudo-labels, we propose to select target pseudo-labels and assign pixels with different weights through an uncertainty-aware pseudo-label learning strategy (UPL) that contains the following three major steps.
%, \textit{i.e.}, 1) estimate the uncertainty of each target pseudo-label by evaluating the consistency among multiple saliency predictions under different data augmentations, 2) select target samples according to the rank of image-level uncertainty, 3) re-weight pixels of the selected pseudo-labels according to the estimated pixel-level uncertainty. 

\begin{figure*}[ht]
    \centering
    \includegraphics[width=\textwidth]{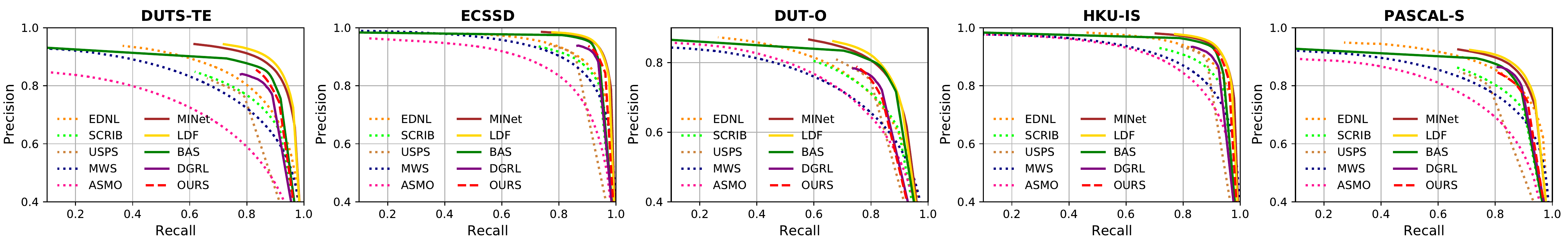}
    %\vspace{-5mm}
    %\subfigure{\includegraphics[width=\textwidth]{figures/f_curve.pdf}}
    %\caption{Quantitative comparison with state-of-the-art SOD methods in terms of Precision-Recall curves (1st row) and F-measure curves (2nd row) on five benchmark SOD datasets.}
    \caption{Quantitative comparison with state-of-the-art SOD methods in terms of Precision-Recall curves.} %on five benchmark SOD datasets.}
    \label{fig:curve}
\end{figure*}

\textbf{1) Consistency-Based Uncertainty Estimation.}
To update the target pseudo-labels, we first perform consistency-based uncertainty estimation. Specifically, as shown in Fig.~\ref{fig:pipeline}, given a saliency detector with fixed parameters $\tilde\theta$, we feed each real target image $I_t$ into the saliency detector to obtain its pseudo-label $\hat y_t=p_{\tilde\theta}(I_t)$. 
To model the uncertainty of a target pseudo-label, we consider the following two aspects. First, the saliency detector will be robust to different small noises on target samples of high-confidence / low-uncertainty. Second, as it is recognized data augmentation can be regarded as a noise injection method~\cite{NEURIPS2020_44feb009}, we model the uncertainty by evaluating the consistency of the saliency predictions of the target image $I_t$ under multiple data augmentations. The salient prediction under the data augmentation $\{\alpha_j(.)\}_{j=1}^N$ can be formulated as:
\begin{equation}
    \tilde y^j_t=\alpha^{-1}_j(p_{\tilde\theta}(\alpha_j(I_t))).
\end{equation}
Here, we only adopt the data augmentation $\alpha(.)$ that can be reversed and $\alpha^{-1}(.)$ will be applied for each saliency prediction $\tilde y^j_t$ to transform it back to the same condition (\textit{e.g.}, direction, scale) as the pseudo-label $\hat y_t$. Inspired by \cite{zheng2021rectifying}, we leverage variance to evaluate the consistency of the pseudo-label and other saliency predictions of different variants of data augmentations. For simplification, we let $\tilde y_t^1=\hat y_t$. The variance map $v_t$ of the sample $I_t$ can be formulated as:
\begin{equation}
    {Var}(I_t, \tilde\theta)=\mathbb{E}[(\tilde y^j_t -\frac{1}{N}(\sum_{j=1}^N\tilde y_t^j))^2], \label{eq:var}
\end{equation}
where $\mathbb{E}(.)$ denotes the mathematical expectation. The dense variance map $v_t \in \mathbb{R}^{H\times W}$ can be used to represent the pixel-level uncertainty of the target pseudo-label $\hat y_t$.

\textbf{2) Image-level Sample Selection (ISS).}
Since the saliency detector is generally weak in the early training stage and is gradually improved during iterative training, we propose that 1) only the pseudo-labels of low uncertainty should be selected and 2) the number of pseudo-label should slowly increase with the increase of training rounds. As shown in Fig.~\ref{fig:pipeline}, the variance maps can reflect the pixel-level uncertainty of the target pseudo-labels, where red and blue indicate high and low uncertainty, respectively. Thus, to rank the target sample by their uncertainty, we introduce the image-level uncertainty score $U$ based on the mean value of variance (Eq.~(\ref{eq:var})). The uncertainty score of the target image $I_t$ can be formulated as:
\begin{equation}
    U(I_t, \tilde\theta) = \frac{1}{HW}\sum_{h=1}^{H}\sum_{w=1}^{W}Var(I_t, \tilde\theta)^{(h,w)}.
\end{equation}
We rank all the target domain samples according to the uncertain score and select a certain proportion of target samples with low uncertainty for each round. The proportion will increase with the improvement of the saliency detector. Note that here we also empirically discard those pseudo-labels composed of nearly all salient or non-salient pixels.

\textbf{3) Pixel-wise Pseudo-Label Reweighting (PPR).}
Although the selected target pseudo-labels generally reflect a low-uncertainty level, there still exists high uncertainty regions
such as object boundaries as shown in their variance maps. Therefore, we suggest that each pixel of the pseudo-labels should be treated differently during the training process and further propose a pixel-wise pseudo-label reweighting strategy $\Omega$ based on the variance maps $Var$. The pixel-wise weight matrix $ w_t\in(0,1]^{H\times W}$ mentioned in Eq.~(\ref{eq:loss_trg}) can be replaced by $\Omega(I_t, \tilde\theta)$ that is computed as:
\begin{equation}
     \Omega(I_t, \tilde\theta)=\exp(-k~Var(I_t, \tilde\theta)),
\end{equation}
where $k\in \mathbb{R}^{+}$ indicates the descent degree of the soft weights. We set $k=20$ in our experiments.

%\subsection{Framework Details}
%\label{sec:details}
%\textbf{Parameters Settings.} 
%We adopt LDF~\cite{wei2020label} as our saliency detector, which uses ResNet-50~\cite{he2016deep} as backbone network. During training, we adopt our proposed SYNSOD (11,197 images) as the source domain dataset and the training set of DUTS~\cite{wang2017learning} (10,533 images) as the target domain dataset. 
%We set the total number of training rounds to 6. Each round has 9K training iterations except that the first round has 12K. 
%The proportion of the selected source and target domain samples are set to \{1.0, 0.5, 0.25, 0.125, 0.0625, 0.03125\} and \{0.0, 0.1, 0.2, 0.4, 0.6, 0.6\} respectively in the 6 rounds. The source samples are randomly selected while the target samples are selected via the proposed image-level sample selection strategy mentioned in Sec.~\ref{sec:upl}. As shown in Fig.~\ref{fig:pipeline}, we use three kinds of data augmentations for uncertainty estimation, including 1) horizontal flip (Flip), 2) rescale input image to $224\times224$ (Scale), and 3) randomly swap image style with other target images via FDA~\cite{yang2020fda}.

%------------------------------------------------------------------------

\section{Experiments}
\subsection{Experimental Setup}
\label{sec:details}
\textbf{Implementation Details.} %Our proposed method is implemented on PyTorch~\cite{paszke2017automatic}, a flexible deep learning framework. 
We adopt ResNet-50-based~\cite{he2016deep} LDF~\cite{wei2020label} as our saliency detector. During training, we adopt SYNSOD (11,197 images) as the source domain and the training set of DUTS~\cite{wang2017learning} (10,533 images) as the target domain. We set the total number of training rounds to six. The proportion of the selected source and target domain samples are set to \{1.0, 0.5, 0.25, 0.125, 0.0625, 0.03125\} and \{0.0, 0.1, 0.2, 0.4, 0.6, 0.6\} respectively in the six rounds. The source samples are randomly selected while the target samples are selected via the proposed image-level sample selection strategy ISS.
%We use an SGD optimizer with momentum of 0.9 and weight decay of 0.0005. We adopt the linear one cycle learning rate policy~\cite{smith2019super} to schedule the learning of each training round and set the maximum learning rate of round $i\in$\{0,1,2,3,4,5\} to $0.005 * 0.75 ^ i$ for ResNet-50 backbone and $0.05 * 0.75 ^ i$ for other parts.
We use an SGD optimizer and adopt the linear one cycle learning rate policy~\cite{smith2019super} to schedule each training round. The whole training process takes about 20 hours with a batch size of 32 on a workstation with a NVDIA GTX 1080 GPU. During testing, each image is resized to $352\times352$, and fed into the network for saliency prediction without any post-processing. More implementation details are provided in the supplemental materials. 

% ---- Table 1 ----- Quantitative Comparison
\begin{table*}
\centering
\resizebox{\textwidth}{!}{%
\begin{tabular}{l|c|ccc|ccc|ccc|ccc|ccc|ccc}
\hline
\multicolumn{1}{l|}{\multirow{2}{*}{Method}} & \multirow{2}{*}{Sup.} & \multicolumn{3}{c|}{DUTS-TE}            & \multicolumn{3}{c|}{ECSSD}           & \multicolumn{3}{c|}{DUT-O}          & \multicolumn{3}{c|}{HKU-IS}          & \multicolumn{3}{c|}{PASCAL-S}        & \multicolumn{3}{c}{SOD}             \\
\multicolumn{1}{l|}{}                        &                       & $S_{m}\uparrow$            & $F_{\beta}^{w}\uparrow$           & $\mathcal{M}\downarrow$           & $S_{m}\uparrow$            & $F_{\beta}^{w}\uparrow$           & $\mathcal{M}\downarrow$           & $S_{m}\uparrow$            & $F_{\beta}^{w}\uparrow$           & $\mathcal{M}\downarrow$           & $S_{m}\uparrow$            & $F_{\beta}^{w}\uparrow$           & $\mathcal{M}\downarrow$           & $S_{m}\uparrow$            & $F_{\beta}^{w}\uparrow$           & $\mathcal{M}\downarrow$           & $S_{m}\uparrow$            & $F_{\beta}^{w}\uparrow$           & $\mathcal{M}\downarrow$           \\ \hline
R3Net~\cite{deng2018r3net}                                           & F\&D                  & .836          & .713          & .066          & .903          & .860          & .056          & .818          & .679          & .071          & .892          & .833          & .048          & .809          & .730          & .104          & .738          & .700          & .136          \\
DGRL~\cite{wang2018detect}                                         & F\&D                  & .842          & .774          & .050          & .903          & .891          & .041          & .806          & .709          & .062          & .894          & .875          & .036          & .836          & .800          & .072          & .774          & .738          & .103          \\
Capsal~\cite{zhang2019capsal}                                       & F\&D                  & .819          & .689          & .063          & .828          & .775          & .073          & .677          & .489          & .099          & .852          & .780          & .058          & .838          & .790          & .073          & .684          & .573          & .152          \\
TSPOA~\cite{liu2019employing}                                        & F\&D                  & .860          & .767          & .049          & .907          & .876          & .046          & .818          & .697          & .061          & .902          & .862          & .038          & .841          & .779          & .078          & .775          & .718          & .115          \\
BASNet~\cite{wang2018detect}                                          & F\&D                  & .866          & .803          & .048          & .916          & .904          & .037          & .836          & .751          & .056          & .909          & .889          & .032          & .836          & .795          & .077          & .772          & .728          & .112          \\
MINet~\cite{pang2020multi}                                     & F\&D                  & .884          & .825          & .037          & \textbf{.925} & .911          & \textbf{.033} & .833          & .738          & .056          & .919          & .897          & .029          & .856          & .814          & .064          & \textbf{.805} & \textbf{.768} & \textbf{.092} \\
GateNet~\cite{zhao2020suppress}                                      & F\&D                  & .885          & .809          & .040          & .920          & .894          & .040          & .838          & .729          & .055          & .915          & .880          & .033          & .858          & .801          & .069          & .801          & .753          & .098          \\
LDF~\cite{wei2020label}                                          & F\&D                  & \textbf{.892} & \textbf{.845} & \textbf{.034} & .924          & \textbf{.915} & .034          & \textbf{.839} & \textbf{.752} & \textbf{.052} & \textbf{.919} & \textbf{.904} & \textbf{.028} & \textbf{.862} & \textbf{.826} & \textbf{.061} & .800          & .765          & .093          \\ \hline
MB+~\cite{zhang2015minimum}                                          & U\&H                  & .595          & .307          & .149          & .595          & .389          & .199          & .612          & .331          & .143          & .609          & .383          & .166          & .528          & .296          & .224          & .490          & .280          & .255          \\
RBD~\cite{zhu2014saliency}                                          & U\&H                  & .567          & .278          & .305          & .667          & .423          & .271          & .572          & .288          & .310          & .648          & .385          & .271          & .621          & .389          & .297          & .612          & .398          & .305          \\
ASMO~\cite{li2018weakly}                                         & W\&D                  & .697          & .488          & .116          & .802          & .702          & .110          & .752          & .559          & .101          & .804          & .701          & .086          & .714          & .578          & .152          & .669          & .551          & .185          \\
%MNL~\cite{zhang2020weakly}                                          & U\&D                  & -             & -             & -             & .870          & .823          & .069          & .788          & .633          & .076          & .884          & .834          & .047          &\textbf{.824}          & .743          & .093          & .736          & .663          & .144          \\
MWS~\cite{zeng2019multi}                                          & W\&D                  & .759          & .586          & .091          & .828          & .716          & .096          & .756          & .527          & .109          & .818          & .685          & .084          & .767          & .614          & .134          & .702          & .571          & .166          \\
USPS~\cite{nguyen2019deepusps}                                         & U\&D                  & .788          & .700          & .068          & .862          & .844          & .062          & .793          & .698          & .063          & .876          & .857          & .041          & .773          & .715          & .108          & .713          & .659          & .143          \\
EDNL~\cite{zhang2020learning}                                         & U\&D                  & .820          & .701          & .065          & .871          & .827          & .068          & .783          & .633          & .076          & .884          & .838          & .046          & .819          & .739          & .095          & .739          & .669          & .142          \\
SCRIB~\cite{zhang2020weakly}                                        & W\&D                  & .803          & .709          & .062          & .865          & .835          & .059          & .785          & .669          & .068          & .865          & .831          & .047          & .796          & .736          & .094          & .727          & .668          & .129          \\ \hline
Ours                                         & U\&D                  & \textbf{.846} & \textbf{.783} & \textbf{.050} & \textbf{.899} & \textbf{.885} & \textbf{.043} & \textbf{.808} & \textbf{.711} & \textbf{.059} & \textbf{.897} & \textbf{.879} & \textbf{.035} & \textbf{.822} & \textbf{.773} & \textbf{.080} & \textbf{.788} & \textbf{.750} & \textbf{.095} \\ \hline
\end{tabular}}
%\vspace{-1mm}
\caption{Quantitative comparison with state-of-the-art SOD methods on six datasets in terms of S-measure $S_m\uparrow$, weighted F-measure $F_\beta^w\uparrow$, and MAE $\mathcal{M}\downarrow$. $\uparrow$ and $\downarrow$ indicate larger and smaller is better, respectively. 
%`-' means the author does not provide corresponding saliency maps. 
The best performance of fully-supervised and weak-/un-supervised methods is marked in \textbf{bold}, respectively. `Sup.' denotes supervision type. `F\&D' means fully-supervised and deep learning-based methods. `U\&H' means unsupervised and handcrafted methods. `W\&D' refers to weakly-supervised and deep learning-based methods. `U\&D' means unsupervised and deep learning-based methods.}
\label{tab:cmp_quantitative}
\end{table*}

\begin{figure*}[ht]
    \centerline{
      \includegraphics[width=0.96\textwidth]{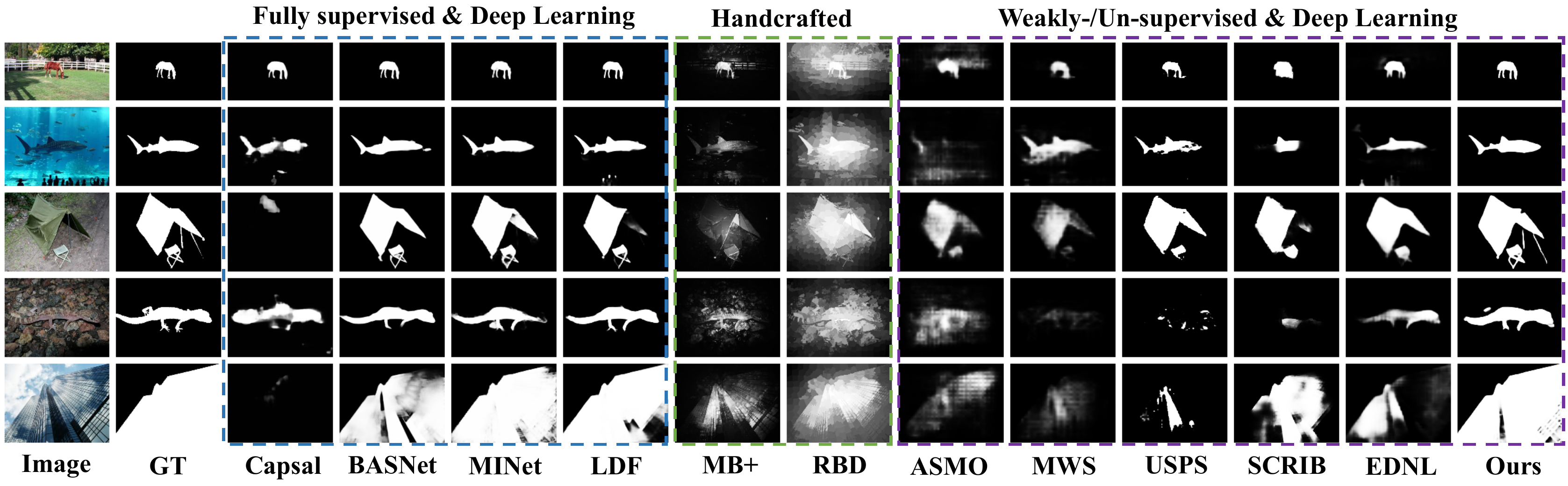}
    }
    %\vspace{-3mm}
    \caption{Visual comparisons of different types of SOD methods, where each row displays an input image. %, the ground truth (GT) saliency map, and the predicted saliency maps generated by four fully supervised methods: Capsal, BASNet, MINet, LDF, two handcrafted methods: MB+, RBD, three weakly supervised methods: ASMO, MWS, SCRIB, three unsupervised learning based  methods: USPS, EDNL, and our proposed method (Ours). 
    Our proposed method~(Ours) consistently generates saliency maps close to the ground truth~(GT).}
    \label{fig:cmp_visual}
\end{figure*}

% Table 2 Ablation
\begin{table*}[ht]
\centering
\resizebox{0.95\textwidth}{!}{%
\begin{tabular}{l|cc|cc|ccc|ccc|ccc|ccc|ccc}
\hline
\multirow{2}{*}{Method} & \multicolumn{2}{c|}{Training} & \multicolumn{2}{c|}{UPL} & \multicolumn{3}{c|}{DUTS-TE~\cite{wang2017learning}} & \multicolumn{3}{c|}{ECSSD~\cite{yan2013hierarchical}} & \multicolumn{3}{c|}{DUT-O~\cite{yang2013saliency}} & \multicolumn{3}{c|}{HKU-IS~\cite{li2015visual}} & \multicolumn{3}{c}{PASCAL-S~\cite{li2014secrets}} \\
            & Source  & Target  & ISS     & PPR     & $S_{m}\uparrow$ & $F_{\beta}^{w}\uparrow$ & $\mathcal{M}\downarrow$ & $S_{m}\uparrow$ & $F_{\beta}^{w}\uparrow$ & $\mathcal{M}\downarrow$ & $S_{m}\uparrow$ & $F_{\beta}^{w}\uparrow$ & $\mathcal{M}\downarrow$ & $S_{m}\uparrow$ & $F_{\beta}^{w}\uparrow$ & $\mathcal{M}\downarrow$ & $S_{m}\uparrow$ & $F_{\beta}^{w}\uparrow$ & $\mathcal{M}\downarrow$ \\ \hline
Source only & $\surd$ &         &         &         & .802            & .695                    & .066            & .873            & .836                    & .060            & .752            & .608                    & .079            & .863            & .814                    & .056            & .795            & .725                    & .103            \\
Vanilla PL  & $\surd$ & $\surd$ &         &         & .818            & .724                    & .067            & .877            & .845                    & .057            & .769            & .643                    & .084            & .875            & .836                    & .049            & .800            & .732                    & .100            \\
UPL w/o ISS & $\surd$ & $\surd$ &         & $\surd$ & .823            & .741                    & .063            & .883            & .861                    & .052            & .778            & .666                    & .077            & .880            & .850                    & .044            & .805            & .749                    & .092            \\
UPL w/o PPR & $\surd$ & $\surd$ & $\surd$ &         & .842            & .777                    & .052            & .894            & .878                    & .046            & .803            & .704                    & .064            & .894            & .875                    & .037            & .822            & .774                    & .082            \\
UPL (Ours)  & $\surd$ & $\surd$ & $\surd$ & $\surd$ & \textbf{.846}            & \textbf{.783}                    & \textbf{.050}            & \textbf{.899}            & \textbf{.885}                    & \textbf{.043}            & \textbf{.808}            & \textbf{.711}                    & \textbf{.059}            & \textbf{.897}            & \textbf{.879}                    & \textbf{.035}            & \textbf{.822}            & \textbf{.774}                    & \textbf{.080}            \\ \hline
\end{tabular}}
%\vspace{-1mm}
\caption{Ablation study on five benchmark datasets using S-measure $S_m\uparrow$, weighted F-measure $F_\beta^w\uparrow$, and MAE $\mathcal{M}\downarrow$.}
\label{tab:ablation}
%\vspace{-3mm}
\end{table*}

\textbf{Datasets and Evaluation Metrics.} To evaluate the performance of our method, we conduct testing on six real-world benchmark SOD datasets including DUTS-TE~\cite{wang2017learning} (5,017 images), ECSSD~\cite{yan2013hierarchical} (1,000 images), DUT-O~\cite{yang2013saliency} (5,168 images), HKU-IS~\cite{li2015visual} (4,447 images), PASCAL-S~\cite{li2014secrets} (850 images), SOD~\cite{movahedi2010design} (300 images). 
%Specifically, DUTS including 10553 training images (\textit{i.e.}, DUTS-TR) and 5019 test images (\textit{i.e.}, DUTS-TE), which is currently largest salient object detection dataset. These images contain complicated scenes.
%ECSSD has 1000 images with semantically meaning and complex contents. Each image has pixel-wise annotations.
%DUT-O is composed by 5168 challenging images with complicated background. HKU-IS includes 4447 challenging images which contain multiple disconnected salient objects. PASCAL-S has 850 image of complex background and high content variety. SOD has 300 challenging images, each of which contains multiple salient object with low contrast.
%\textbf{Evaluation Metrics.} 
We adopt four widely used evaluation metrics, \textit{i.e.}, precision-recall (PR) curve, mean absolute error (MAE, $\mathcal{M}$)~\cite{perazzi2012saliency}, weighted F-measure ($F_\beta^w$) ~\cite{margolin2014evaluate}, and S-measure ($S_m$)~\cite{fan2017structure}.
%The pairs of precision and recall are calculated by comparing the ground truth saliency map with the saliency map binarized using different thresholds( [0, 255]). The PR curve of a dataset can be obtained from all pairs of average precision and recall values over all saliency maps.
%\textit{F-measure} indicates the performance at different thresholds, which synthetically considers both precision and recall: $F_{\beta} = \frac{(1 + \beta^2) \cdot precision \cdot recall}{ \beta^2  \cdot precision +  recall},$ where $\beta$ is set to 0.3 as suggested in \cite{achanta2009frequency} to weight precision more than recall. We report the whole F-measure curve in this paper. 
%MAE computes the average per-pixel absolute error between normalized predicted saliency map and ground truth. 
%Weighted F-measure is weighted region-based similarity metric that defined as $F_{\beta}^w = \frac{(1 + \beta^2) \cdot precision^w \cdot recall^w}{ \beta^2  \cdot precision^w +  recall^w},$ where $\beta$ is set to 0.3 as suggested in \cite{achanta2009frequency}.
%S-measure measures structural similarity between the saliency map and ground truth. 
%It considers both region-aware ($S_r$) and object-aware ($S_o$) similarities: $S_m = \alpha * S_o + (1 - \alpha)  * S_r,$ where $\alpha$ is set to 0.5 as suggested by \cite{fan2017structure}.

\subsection{Comparison with State-of-the-Art}

\textbf{Quantitative Comparison.} In Table~\ref{tab:cmp_quantitative}, we compare our method with eight fully supervised deep saliency prediction methods: R3Net, DGRL, Capsal, TSPOA, BASNet, MINet, GateNet, LDF, two handcrafted unsupervised methods : MB+, RBD, and five deep weakly-/un-supervised methods: ASMO, MWS, SCRIB, USPS, EDNL. 
For a fair comparison, we evaluate all the saliency maps provided by the authors with the same evaluation code. 
%We presents the quantitative comparison in terms of S-measure $S_m$, weighted F-measure $F_\beta^w$, and MAE $\mathcal{M}$. 
As shown in the table, our method consistently outperforms existing weakly-supervised and unsupervised SOD methods by a large margin over all six datasets. Specifically our method achieves an average gain of 3.65\%, 5.56\% 1.61\% w.r.t $S_{m}$, $F_{\beta}^{w}$ and $\mathcal{M}$ compared with previous state-of-the-art weakly-supervised method SCRIB~\cite{zhang2020weakly} on six datasets. As for previous state-of-the-art deep unsupervised method EDNL~\cite{zhang2020learning}, our approach obtains an average gain of 2.4\%, 6,23\%, 2.15\% w.r.t $S_{m}$, $F_{\beta}^{w}$ and $\mathcal{M}$ over six datasets. Moreover, the performance of our proposed UDASOD method is comparable to state-of-the-art fully-supervised SOD methods, and even better than several of them, such as R3Net~\cite{deng2018r3net}, DGRL~\cite{wang2018detect}, TSPOA~\cite{liu2019employing}.
Fig.~\ref{fig:curve} presents the precision-recall curves of different SOD methods on five datasets, where weakly-/un-supervised methods are represented by dotted lines. From the figure, we can observe that our method overall lies above other weakly-/un-supervised methods and is even comparable to some fully supervised methods.%, which verifies the superiority of our proposed method.

\textbf{Qualitative Comparison.} Fig.~\ref{fig:cmp_visual} presents several representative visual examples of predicted saliency maps. These examples reflect various scenarios, including small object (1st row), object with a complex background (2nd row), object with thread-like boundary (3rd row), low contrast between salient object and image background (4th row), and object with a border-connected region (5th row). It can be seen that our proposed method produces accurate and complete saliency maps with sharp boundaries and coherent details, which consistently outperforms the weakly-/un-supervised models and even some fully supervised models. 

% Table 3
\begin{table}
\centering
\resizebox{0.43\textwidth}{!}{%
\begin{tabular}{l|cc|cc|cc|cc}
\hline
\multicolumn{1}{l|}{\multirow{2}{*}{Augmentation}} & \multicolumn{2}{c|}{DUTS-TE}                  & \multicolumn{2}{c|}{ECSSD}                 & \multicolumn{2}{c|}{DUT-O}                 & \multicolumn{2}{c}{HKU-IS}                \\
\multicolumn{1}{l|}{}                              & $F_{\beta}^{w}\uparrow$ & $\mathcal{M}\downarrow$ & $F_{\beta}^{w}\uparrow$ & $\mathcal{M}\downarrow$ & $F_{\beta}^{w}\uparrow$ & $\mathcal{M}\downarrow$ & $F_{\beta}^{w}\uparrow$ & $\mathcal{M}\downarrow$ \\ \hline
Scale & .765                    & .055            & .873                    & .049            & .664                    & .072            & .869                    & .039            \\
FDA                                                & .774                    & .053            & .872                    & .047            & .700                    & .067            & .874                    & .038            \\
Flip & .777                    & .053            & .879                    & .045            & .697                    & .066            & .875                    & .038            \\
Flip+Scale+FDA                                 & \textbf{.783}                    & \textbf{.050}            & \textbf{.885}                    & \textbf{.043}            & \textbf{.711}                    & \textbf{.059}            & \textbf{.879}                    & \textbf{.035}            \\ \hline
\end{tabular}}
%\vspace{-1mm}
\caption{Sensitivity to different kinds of data augmentation in consistency-based uncertainty estimation.}
\label{tab:aug}
\end{table}

\subsection{Ablation Study}
\textbf{Effectiveness of UDASOD.}
To demonstrate the effectiveness of our proposed unsupervised domain adaptive salient object detection (UDASOD) through the uncertainty-aware pseudo-label learning (UPL) strategy, we conduct the ablation study from the following aspects and report the performance of different variants in Table~\ref{tab:ablation}.

\noindent\textbf{1) Synthetic Data.} 
The saliency detector trained with only synthetic source data (Source only) achieves comparable performance to other unsupervised models (as shown in Table~\ref{tab:cmp_quantitative}), indicating the feasibility of learning salient object detection from the proposed synthetic dataset SYNSOD.

\noindent\textbf{2) Unsupervised Domain Adaption.}
Introducing the unlabeled real target data through vanilla pseudo-label learning (Vanilla PL) strategy can improve the performance of source only model, which demonstrates that a simple unsupervised domain adaption through pseudo-label learning can help to mitigate the domain gap between the synthetic and real domains. While our proposed method (UPL) can further boost the performance of vanilla PL by a large margin by exploiting image-level sample 
selection (ISS) and pixel-level pseudo-label reweighting (PPR).

\noindent\textbf{3) Uncertainty-Aware Pseudo-Label Learning.} 
To further verify the effectiveness of each component in the proposed UPL. We conduct the ablation study by removing PPR and ISS from UPL, respectively, \textit{i.e.}, UPL w/o PPR and ISS w/o ISS in Table~\ref{tab:ablation}. 
Compared to UPL, the performance of UPL w/o PPR slightly drops on five datasets, which indicates that the selected low uncertainty pseudo-labels still contain some misclassified pixels and the PPR module can alleviate the noise of pseudo labels by adjusting the weights of pixels. 
UPL w/o ISS is iteractively trained with all the target pseudo-labels without image-level selection, resulting in a severe performance degradation compared to UPL. Theoretically, image-level selection can be approximated as a special case of pixel-level reweighting. However, in practice, using only pixel-level reweighting (UPL w/o ISS) performs worse than image-level selection (UPL w/o PPR). We conjecture that without image-level selection, the pseudo-labels of those high-uncertainty samples naturally have lots of misclassified pixels that will be suppressed by the pixel-wise reweighting. As suggested by\cite{shin2020two}, this will lead to sparse pseudo-labels and inevitably increase the difficulty of network convergence. Whereas, the ISS and PPR modules are complementary to each other and can further boost the performance of our proposed method. 

\begin{figure}[t]
    \centerline{
      \includegraphics[width=0.43\textwidth]{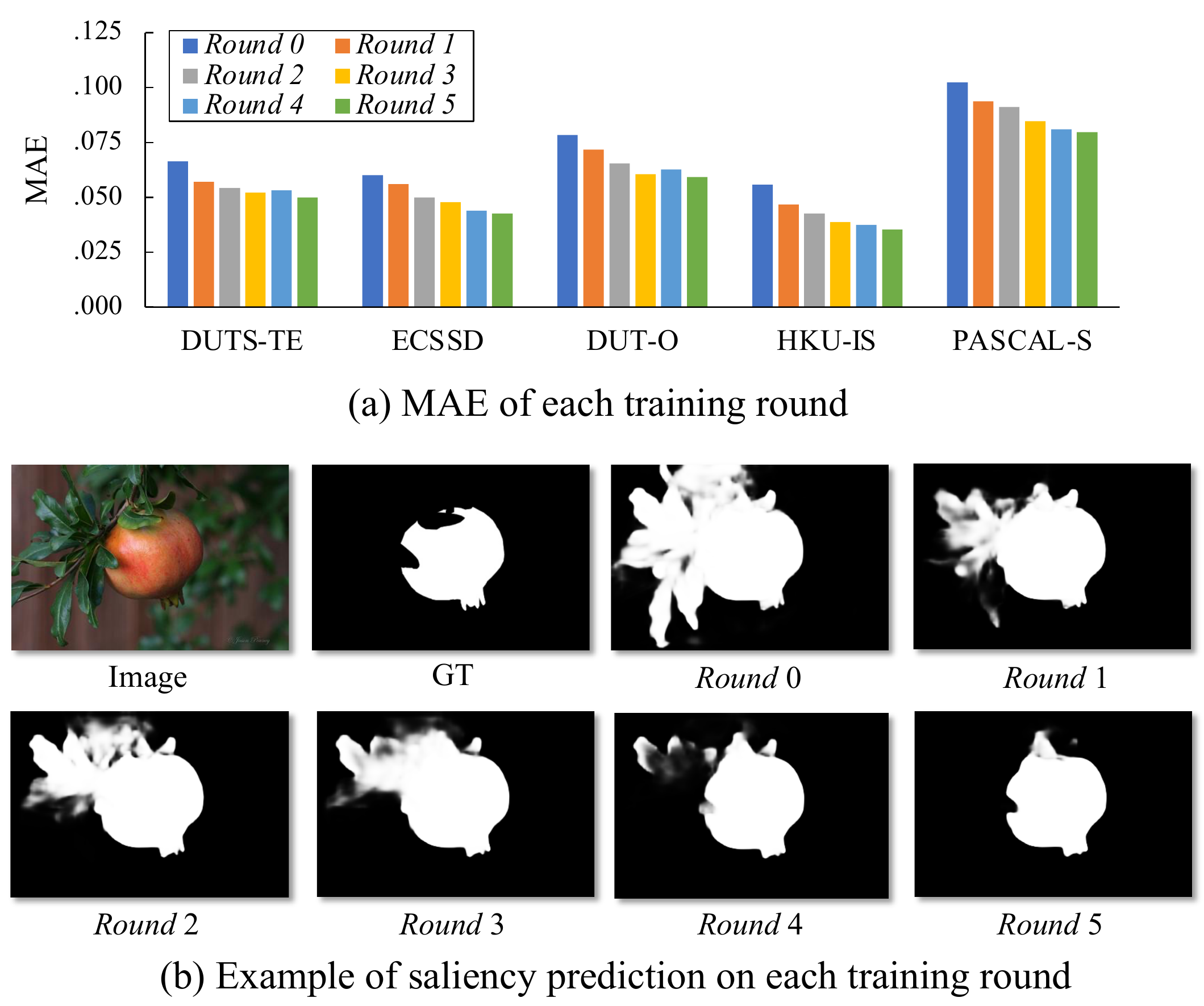}
    }
    \caption{Quantitative and visual performance of our proposed method on each training round.}
    %\vspace{-1mm}
    \label{fig:cmp_round}
\end{figure}

\textbf{Sensitivity to Data Augmentation.}
Our proposed method leverages multiple data augmentations as noise injection methods to estimate the uncertainty of pseudo-labels. To demonstrate that our method is applicable to different data augmentations, we report the performance using the augmentations mentioned in UPL. As shown in Table~\ref{tab:aug}, our proposed method is not limited to a single kind of data augmentation. When applying only one data augmentation (\textit{i.e.}, Flip, Scale, FDA) the proposed uncertainty-aware pseudo-label learning (UPL) strategy can still work and outperform the vanilla pseudo-label learning strategy (Vanilla PL in Table~\ref{tab:ablation}) by a large margin, which indicates the robustness of our proposed UPL. Moreover, when combining different data augmentations (Flip+Scale+FDA), the performance of UPL can be further improved as the combination leads to more stable uncertainty measurement. 

\textbf{Sensitivity to Training Rounds.}
Our proposed method adopts an iterative training paradigm that contains multiple rounds. To show the performance of each training round more intuitively, we present the MAE results and predicted saliency maps in Fig.~\ref{fig:cmp_round}. As shown in Fig.~\ref{fig:cmp_round} (a), MAE is consistently improved with the increase of training rounds over all datasets. Moreover, as shown in Fig.~\ref{fig:cmp_round} (b), the non-salient pixels of the predicted saliency map are gradually suppressed and lead to a more accurate result.

%------------------------------------------------------------------------
\section{Conclusion}
In this paper, we propose to tackle deep unsupervised salient object detection from a novel perspective, \textit{i.e.}, learning from synthetic but clean labels. To achieve this goal, we construct a new synthetic salient object detection dataset and introduce a novel unsupervised domain adaptive salient object detection framework to learn and adapt from the synthetic dataset. Specifically, the proposed algorithm exploiting an uncertainty-aware pseudo-label learning strategy to mitigate the domain gap between the synthetic source domain and the real target domain. Extensive experiments on multiple benchmark datasets demonstrate the effectiveness and robustness of our proposed method, which makes it superior to all state-of-the-art deep unsupervised methods and even comparable to fully-supervised methods.

\section{Acknowledgements}
This work was supported in part by the Chinese Key-Area Research and Development Program of Guangdong Province (2020B0101350001), in part by the Guangdong Basic and Applied Basic Research Foundation under Grant No.2020B1515020048, in part by the National Natural Science Foundation of China under Grant No.61976250, No.U1811463 and No.61906049, and in part by the Guangdong Provincial Key Laboratory of Big Data Computing, the Chinese University of Hong Kong, Shenzhen.

{\small
\bibliography{udasod}
}

\end{document}